\documentclass{article}

\usepackage{microtype}
\usepackage{graphicx}
\usepackage{subfigure}
\usepackage{booktabs}

\usepackage{hyperref}

\usepackage[accepted]{icml2024}

\usepackage{amsmath}
\usepackage{amssymb}
\usepackage{mathtools}
\usepackage{amsthm}
\usepackage{xfrac}
\usepackage{bm}
\usepackage{yhmath}

\usepackage[most]{tcolorbox}
\tcbset{size=small, frame hidden, colback=white!98!blue!98!green, colframe=white!85!blue!95!green, grow to left by=0.2cm, grow to right by=0.2cm}

\usepackage[capitalize,noabbrev]{cleveref}

\theoremstyle{plain}

\theoremstyle{definition}

\theoremstyle{remark}

\usepackage[textsize=tiny,disable]{todonotes}

\icmltitlerunning{What makes an image realistic?}

\DeclarePairedDelimiterX{\infdivx}[2]{[}{]}{%
  #1\delimsize\| #2%
}

\newcommand{\KL}{D_{\mathrm{KL}}}
\newcommand{\KLD}{\KL\infdivx}

\newcommand{\x}{{\mathbf{x}}}

\newcommand{\y}{{\mathbf{y}}}
\newcommand{\z}{{\mathbf{z}}}

\begin{document}

\twocolumn[
  \icmltitle{What makes an image realistic?}

\icmlsetsymbol{equal}{*}

\begin{icmlauthorlist}
\icmlauthor{Lucas Theis}{dm}
\end{icmlauthorlist}

\icmlaffiliation{dm}{Google DeepMind, London, UK}

\icmlcorrespondingauthor{Lucas Theis}{theis@google.com}

\icmlkeywords{perceptual quality, realism, neural compression, generative adversarial networks}

\vskip 0.3in
]

\printAffiliationsAndNotice{} 

\begin{abstract}
The last decade has seen tremendous progress in our ability to \textit{generate} realistic-looking data, be it images, text, audio, or video. Here, we discuss the closely related problem of \textit{quantifying} realism, that is, designing functions that can reliably tell realistic data from unrealistic data. This problem turns out to be significantly harder to solve and remains poorly understood, despite its prevalence in machine learning and recent breakthroughs in generative AI. Drawing on insights from algorithmic information theory, we discuss why this problem is challenging, why a good generative model alone is insufficient to solve it, and what a good solution would look like. In particular, we introduce the notion of a \textit{universal critic}, which unlike adversarial critics does not require adversarial training. While universal critics are not immediately practical, they can serve both as a North Star for guiding practical implementations and as a tool for analyzing existing attempts to capture realism.
\end{abstract}

\section{Introduction}
\label{sec:intro}

What distinguishes realistic images from unrealistic images? Humans are able to detect a wide variety of flaws in images and other sensory data, yet there are no robust losses which could be used to penalize unrealistic images across a broad set of tasks in machine learning, and no widely accepted formal notion of realism exists today. In particular, we are interested in real-valued functions $U$ producing a low value $U(\x)$ when some data $\x$ is \textit{realistic} and a large value when $\x$ is \textit{unrealistic}. Here, $\x$ could be a single image, a small set of images, or a video. But our discussion will also be relevant for other types of data such as text of arbitrary length or more generally any data drawn from some distribution which we will denote~$P$. 

Potential applications of such functions are plentiful and include anomaly detection \cite{ruff2021anomaly}, deepfake detection \citep[][]{sha2023defake,pondoc2023seeing}, generative model evaluation \citep[][]{theis2016note,heusel2017fid,borji2019}, model distillation \citep[][]{vandenoor2018distill,yin2023onestep}, neural compression \citep{balle2021ntc,yang2023a}, computational photography \citep[][]{fang2020perceptual}, and computer graphics \citep[][]{herzog2012norm,reinhard2013realism,poole2023df}. Unfortunately, their implementation is extremely challenging. Our ability to \textit{generate} realistic data is rapidly improving \citep[e.g.,][]{dhariwal2021diffusion} yet no reliable candidates or recipes for constructing $U$ exist in machine learning today. This is not for a lack of trying. 
While some progress has been made in the \textit{detection} of unrealistic examples, the design of functions that are robust to \textit{optimization}
(for tasks involving generation) has been less successful.
The latter problem is significantly harder because our function now not only has to detect a limited set of artefacts but has to anticipate any unrealistic examples an optimization might run into. Weaknesses in a function's design often only make themselves known once we start optimizing \cite{ding2021iqa}. Complicating the matter is the fact that the optimization depends on $U$ itself.

To give a more concrete example of the kind of tasks we are interested in, consider the following loss which naturally comes up in lossy compression. If $\x = g(\z)$ is the output of a neural network, we may want to find a representation $\z$ such that
\begin{align}
    R(\z) + \alpha d(\x, \x^*) + \beta U(\x)
    \label{eq:rdp}
\end{align}
is minimal, where $d$ measures the distance to some target image $\x^*$ and $R$ is the number of bits required to encode $\z$.

%So far we have yet to define what we mean by ``realistic''. Intuitively, $\x$ is a realistic example of a distribution $P$ if it is believable that $\x$ has been generated by $P$.
In this paper we will take the view that $\x$ is realistic if it appears to have come about in a particular way, which is another way of saying that $\x$ is a plausible sample of a distribution $P$ capturing the data generating process. What is considered realistic therefore depends on $P$. If $P$ is a distribution over natural images then most photos would qualify as realistic. While an MNIST image \citep{lecun2010mnist} would not be considered a realistic example of a natural image, we would still consider it to be realistic if $P$ is the distibution of MNIST digits.

In Section~\ref{sec:naive} we will first review why common approaches to formalizing realism in terms of probability and typicality fail. This will highlight the challenges involved in defining realism and provide motivation for later sections. In Section~\ref{sec:divergences} we will review much more successful notions of realism based on divergences, adversarial losses, and feature statistics, and discuss how they still fall short of our goal. In Section~\ref{sec:uc} we will make the case that \textbf{\textit{randomness deficiency} \citep{li1997intro} captures realism} and introduce the concept of a \textit{universal critic}. Finally, in Section~\ref{sec:related} we will apply our newly gained understanding of realism to examples from the machine learning literature.
% TODO: remove textbf for arxiv version

What has been referred to as \textit{realism} \citep[e.g.,][]{shaojing2018realism,theis2021coding,careil2023towards} is also often referred to as \textit{perceptual quality} \citep[e.g.,][]{blau2018tradeoff,fang2020perceptual,salehkalaibar2023perception}. It is therefore natural to wonder to what extent human perception should factor into its formalization. Our approach to defining realism is normative, that is, we consider how an idealized observer would judge realism. Similar to how Bayesian inference does not take inspiration from neuroscience but Bayesian decisions resemble human decisions \citep[e.g.,][]{knill2004bayes}, we too can hope that human perception agrees with our definition of realism because it addresses a similar task as that faced by humans.
In Section~\ref{sec:observer}, we will further make the case that \textit{batched universal critics} not only generalize no-reference metrics and divergences---which represent the prevalent ways of formalizing realism---but are also a better model of a human observer.

\section{Probability and Typicality}
\label{sec:naive}

In this section we review the two most common approaches to capture realism found in machine learning, namely those based on probability and typicality, and their failures. Similar failures of probability and typicality have been documented in the anomaly detection literature \citep[e.g.,][]{choi2019waic,charline2021anomaly,osada2023typicality} but are worth repeating as they continue to be a source of confusion.

\subsection{Probability}
\label{sec:probability}

If $\x$ is discrete, it is natural to consider its probability under $P$ to determine whether it is a realistic example of $P$. After all, if $\x$ has low probability then it seems unlikely to have come from $P$. This intuition is widespread in machine learning. Unsupervised anomaly detection, for instance, generally defines anomalies as those data points having low probability or density under a distribution of \textit{normal} examples \citep{ruff2021anomaly}, where the probability is often measured in some feature space \citep[e.g.,][]{zong2018gmm}.
Probability density is also frequently maximized in an attempt to guide synthetic images towards more realistic examples \citep[e.g.,][]{sonderby2016affgan,graikos2022diffusion}.
To see how this approach might fail, consider the following simple example.

\begin{tcolorbox}
    \textbf{Example 1 (Probability).} Consider a computer program simulating a sequence of independent and nearly unbiased coin tosses, $\x^N = (x_1, \dots, x_N)$ with $P(x_n = 1) = 0.5 + \varepsilon$ for some very small $\varepsilon > 0$. For reasonably large $N$, we would expect the program to output a number of 1s which is close to N/2 and we would suspect a bug if the program outputs a sequence of only 1s, yet this is the most probable sequence.
\end{tcolorbox}

Example~1 shows that maximizing $P(\x)$ can lead to unrealistic examples. It also shows that $P(\x)$ would not detect a bug which causes a program to only output 1s.
If instead we count the number of 1s, $k = \sum_{n = 1}^N x_n$, and measure the probability of $k$, this bug could be detected. Does this mean we only need to find the right set of features? By ignoring some aspects of the data, we risk not detecting unrealistic examples. We might therefore conclude that we simply need to test sufficiently many features. Unfortunately this approach also runs into trouble. Consider testing whether $\x$ has 1s in even places and 0s in odd places, $\x = \texttt{0101..01}$. The probability of this sequence is approximately $2^{-N}$ so that we would reject it with high confidence if we happen to observe it. However, since all sequences have roughly the same probability, we would reject \textit{every} sequence as unrealistic if we tested \textit{all} features identifying a specific sequence.

Using densities instead of probabilities introduces an additional challenge, namely that our answer now depends on the parametrization of the data. If $P$ is an exponential distribution with rate 1, say, then values of $\x$ close to zero seem preferable over larger values if judged by their density. But if we consider $\y = e^{-\x}$ instead, then all values of $\y$ would now be considered equally preferable.

\subsection{Weak Typicality}
\label{sec:typicality}

Many readers will not have been surprised by the inability of probabilities to capture realism thanks to the widely known \textit{asymptotic equipartition property} (AEP) of random sequences \citep{cover2006elements}. This property is such that if $\x^N = (\x_1, \dots, \x_N)$ is a sequence of i.i.d. random variables drawn from $P$, then with probability 1 we have
\begin{align}
    \textstyle \lim_{N \rightarrow \infty} -\frac{1}{N} \log P(\x_1, \dots, \x_N) = H[\x_n]
\end{align}
almost surely, where $H[\x_n]$ is the entropy of $P$.
The \textit{typical set} is defined as \citep{cover2006elements}
\begin{align}
    \textstyle A^N_{\delta} = \{ \x : | -\frac{1}{N} \log P(\x^N) - H[\x_n] | < \delta \}
    \label{eq:typical_set}
\end{align}
and elements from this set are considered \textit{weakly typical}. While other notions of typicality exist, weak typicality is the one most commonly encountered in the machine learning literature \citep{nalisnick2019typicality,choi2019waic,dieleman2020typicality}.
The AEP implies that as $N$ increases, the probability that a randomly drawn sequence is contained in the typical set $A^N_{\delta}$ approaches~1 for any $\delta > 0$. That is, a realistic sequence is likely to be typical.
\citep[While we have stated the AEP for sequences of independent and discrete random variables, generalizations to dependent and continuous sources exist and are well known; e.g.,][]{algoet1988aep}.

The above suggests that instead of expecting the probability to be large, we should expect realistic $\x$ to have negative log-probability close to the entropy---or the probability to be roughly $2^{-H[\x]}$, especially if $\x$ is high-dimensional. It therefore appears that $| -\log P(\x) - H[\x] |$ would be a good candidate for a measure of realism \citep{choi2019waic,nalisnick2019typicality}. Unfortunately, also this definition fails to quantify realism as the following examples demonstrate.

\begin{tcolorbox}
\textbf{Example 2 (Typicality).} Consider again a sequence of independent coin tosses. If the coin is unbiased, then the log-probability of any sequence is exactly the entropy, $-\log_2 P(\x^N) = N$. In other words, in this case the probability of $\x^N$ under $P$ is completely uninformative and the typical set contains \textit{every} sequence. Does this mean that every sequence of coin flips is realistic? Clearly, there is a sense in which the sequence \texttt{0000000000} is less realistic than \texttt{1100010100} which is not captured by weak typicality.
\end{tcolorbox}

\begin{tcolorbox}
\textbf{Example 3 (Typicality).} As another example, consider a multivariate Gaussian distribution with density $p(\x) \propto \exp(-\|\x\|^2)$.
With high probability, the negative log-density of a random sample will be close to the differential entropy, which amounts to the norm $\|\x\|$ being roughly constant. While we would expect realistic examples from our distribution to look like uncorrelated noise, optimizing for typicality will only constrain the norm. If $\x$ represents an image, our optimization will merely adjust its contrast but will not decorrelate pixels as one might hope.
\end{tcolorbox}

Weak typicality may be a necessary criterion for realism but it is clearly not sufficient. Put differently, the typical set contains the realistic sequences we care about but also many sequences which are unrealistic, such as long sequences of fair coin flips which all come up heads.

Probability and typicality both fail as a measure of realism because they address the wrong question. They tell us something about $\x$ \textit{assuming that $\x$ has distribution $P$}. 
However, we cannot make this assumption, since whether or not $\x$ follows $P$ is precisely the question we are trying to answer. 
That is, \textbf{we are not interested in the probability (or typicality) of $\x$ given $P$, but in the probability of $P$ given $\x$.}

An extended discussion of typicality can be found in Appendix~\ref{app:typicality}.

\section{Divergences}
\label{sec:divergences}

More successful notions of realism are based on divergences between a ground-truth data distribution $P$ and a distribution $Q$ which we are trying to evaluate. In line with our intuitive notion of realism, if a divergence is zero, then instances of $Q$ are indistinguishable from instances of $P$, that is, we have perfect realism.

In coding theory, formalizing realism in terms of divergences \citep{matsumoto2018rdp,blau2019rethinking,chen2022rdpf} has resulted in an improved understanding of the lossy compression problem and novel methods to solve them \citep[e.g.,][]{theis2021advantages}. In practical applications, generative adversarial networks \citep[GANs;][]{goodfellow2014gan} trained with adversarial losses (which approximate divergences) significantly advanced the state of the art in the perceptual quality of generated images \citep[e.g.,][]{denton2015gan,ledig2017srgan}.
For the evaluation of generated images, the Fréchet inception distance \citep{heusel2017fid} has established itself as the method of choice and is based on a divergence between distributions over feature activations.

In the following, we review two approaches to approximating divergences based on samples.

\subsection{Adversarial Losses}

Adversarial losses provide lower bounds on divergences. For the broad class of $f$-divergences \citep{renyi1961fdiv} between two distributions with densities $p$ and $q$, we can write
\begin{align}
    D_f[q\,\|\,p] = \int p(\x) f\left(\frac{q(\x)}{p(\x)} \right) d\x,
\end{align}
where $f$ is a convex function with $f(1) = 0$. This class of divergences includes the Jensen-Shannon divergence, the total variation distance, and the Kullback-Leibler divergences. For a real-valued function $T$ (with an appropriately limited output range), we obtain the lower bound \citep{nguyen2010estimating,nowozin2016fgan}
\begin{align}
    D_f[q\,\|\,p] \geq \mathbb{E}_q[T(\x)] - \mathbb{E}_p[f^*(T(\x))],
\end{align}
where $f^*$ is the convex conjugate of $f$. $T$ acts as a \textit{critic} whose purpose is to produce values which are large for samples drawn from $q$ and small for samples drawn from $p$. In practice, the critic may be a neural network~$T_{\bm{\theta}}$ and adversarial training amounts to alternating between maximizing the lower bound with respect to its parameters, $\bm{\theta}$, and minimizing the bound with respect to the parameters of $q$ (although practical implementations often deviate from this basic recipe).

For the Kullback-Leibler divergence, for instance, we have
\begin{align}
    f(u) &= u \log u, & f^*(t) = \exp(t - 1),
\end{align}
and the bound is tight for
\begin{align}
    T_q(\x) = \log q(\x) - \log p(\x) + 1.
    \label{eq:optimal_critic}
\end{align}
Note that this optimal critic depends on the distribution $q$ that we are trying to evaluate. In contrast, in our setting we may only have access to a single instance or a few instances drawn from $q$. Furthermore, the dependence of the critic on $q$ is responsible for optimization instabilities that are known to plague adversarial training and which we would like to avoid. In Section~\ref{sec:uc} we will discuss critics which are \textit{universal} in the sense that they do not depend on $q$ and therefore do not require adversarial training.

\subsection{Maximum Mean Discrepancy}
\label{sec:mmd}

\textit{Maximum mean discrepancy} \citep[MMD;][]{gretton2012mmd} refers to a class of divergences which have been used for hypothesis testing as well as for generating realistic images \citep{li2015gmm,dziugaite2015mmd}. Given two sets of i.i.d. examples---$\x_1, \dots, \x_M$ and $\tilde\x_1, \dots, \tilde\x_N$---estimates of MMD can be used to decide whether the two sets were drawn from the same distribution. Formally, we compute
\begin{align*}
   %\widehat{\text{MMD}^2}(\x, \x')
   \text{MMD}^2(\x^M, \tilde\x^N)
   = \textstyle\left\| \frac{1}{M} \sum_m \Phi(\x_m) - \frac{1}{N} \sum_n \Phi(\tilde\x_n) \right\|^2
\end{align*}
in some potentially very high (even infinite) dimensional feature space~$\Phi$ to estimate a squared MMD. Notably, the estimator depends on the two distributions only through examples and unlike adversarial losses does not require optimization of any critic. This makes it worthwhile to consider as a candidate for our function $U$, especially in regimes where we have access to at least a small number of unrealistic examples. The basic idea is that we would fix a relatively large number of realistic examples and compare it to a small batch of examples we wish to test for realism.
Support for this idea also comes from \citet{amir2021mmd} who have shown that MMD can be used to construct an effective full-reference perceptual metric\footnote{A full-reference metric takes two images as arguments where a no-reference metric only has a single input.} which agrees with human judgments in determining the similarity of pairs of images. To construct the metric, each image was treated as a distribution over small patches.

It remains unclear how to use MMD to quantify the realism of a single data point without a reference. For an image, one might compare features averaged over image patches to the features obtained from patches of a larger dataset of images, and similar ideas have shown promise in image quality assessment \citep[e.g.,][]{mittal2013niqe,zhang2015ilniqe}. But the limitations of this approach are also clear as not all realistic images have statistics representative of the entire data distribution.

A bigger concern perhaps is that the statistical power of MMD can drop quickly as the dimensionality of the problem increases\footnote{This assumes that the difficulty of the estimation problem remains constant, as measured by the KL divergence between the two distributions being tested.} \citep{ramdas2015mmd}, suggesting that we might need a very large number of examples if we want to identify defects in reasonably sized images or videos. 

The MMD estimator makes fewer assumptions than is necessary for us. In particular, it seems reasonable to assume access to $P$ (or a good approximation) both from a conceptual and a practical point of view, given the power of today's generative models. By incorporating $P$ into our definition of realism, we can hope to quantify realism more efficiently. MMD leaves it to us to choose $\Phi$ and does not provide a clear mechanism for incorporating~$P$.

\section{Universal Critics}
\label{sec:uc}

In this section, we introduce an alternative notion of realism based on concepts from algorithmic information theory (AIT) \citep{lof1966random,chaitin1987,li1997intro}. AIT is concerned with whether a given sequence of bits is a \textit{random} sequence of independent coin flips. If we can answer this question, then the answer to the more general question of whether $\x$ is an instance of $P$ directly follows, since if we use $P$ to (losslessly) compress $\x$ then the resulting bits should appear random. Several notions of randomness have been proposed and studied in AIT. Some have been rejected on the basis of flaws, such as von Mises randomness \citep{mises1919}. Other notions survived scrutiny and turned out to be equivalent \citep[Chapter 3]{chaitin2001exploring}, namely Martin-Löf randomness \citep{lof1966random}, Solovay randomness \citep{solovay1975}, incompressibility \citep{li1997intro}, and Chaitin randomness \citep{chaitin2001exploring}. The fact that multiple authors converged to essentially the same answer should give us hope that there is something fundamental about the concepts they discovered. Instead of reviewing the different (equivalent) definitions of randomness, we start with the conclusion relevant for us and then develop a justification for it below. In particular, AIT suggests the following measure of randomness to decide whether $\x$ was drawn from a distribution~$P$:
\begin{align}
    U(\x) = -\log P(\x) - K(\x)
    \label{eq:uc}
\end{align}
Here, $K(\x)$ is the \textit{Kolmogorov complexity} of $\x$ which is defined as the length of a shortest program (in some Turing complete programming language) which outputs $\x$. The quantity $U(\x)$ is also known as \textit{randomness deficiency}\footnote{A more accurate definition of randomness deficiency would be over sequences of arbitrary length but for simplicity we will work with Eq.~\ref{eq:uc}.} \citep{li1997intro} but for reasons that will become clear soon, we will refer to $U$ as a \textit{universal critic}. 

The following characterization of Kolmogorov complexity will be more convenient for us,
\begin{align}
    K(\x) = -\log S(\x),
    \quad  
    \textstyle S(\x) = \sum_n \pi_n Q_n(\x),
\end{align}
where $S(\x)$ is \textit{Solomonoff's probability} \citep{solomonoff1960prob} and requires some explanation. Consider the set of all discrete probability distributions implementable in a programming language of your choice. Each program corresponds to a sequence of bits and we are free to interpret those bits as a natural number. In other words, the set of computable probability distributions is countable and we can assign each such distribution $Q_n$ a number $n$. $S$ is a mixture of all of these. The choice of weights $\pi_n$ is not critical for now and we can choose $\pi_n \propto 1/n^2$ or $\pi_n = 2^{-C(n)}$ where $C(n)$ is the number of bits assigned to $n$ by some universal code.

A similar argument holds for continuous sample spaces \citep[Chapter 4.5]{li1997intro}. That is, there is a corresponding $S$ for continuous sample spaces which sums over measures, or lower semicomputable semimeasures\footnote{A semimeasure integrates to a value less or equal 1. $S$ itself is an example of a semimeasure with $\sum_\x S(\x) < 1$. This is due to the \textit{halting problem} causing some unknowable set of indices $n$ to correspond to programs which never stop running. For these $n$, we set $Q_n(\x) = 0$.} to be precise. A measure is \textit{semicomputable} if it can be approximated from below to arbitrary precision, that is, it is enough to be able to compute approximations of a measure for it to be included in the mixture $S$. For simplicity, we will focus on discrete spaces even though continuous spaces are relevant in practice if we want to optimize for realism.

For a more thorough treatment of these concepts, see the excellent introduction to Kolmogorov complexity by \citet{li1997intro}. Here we will try to not get hung up on technical details since we are ultimately interested in practical applications and---as some readers may already rightfully object---Kolmogorov complexity and $S$ are uncomputable.
Nevertheless, we will argue that universal critics as defined in Eq.~\ref{eq:uc} correctly formalize realism, and that it is useful to understand practical approaches as (good or bad) approximations of it---similar to how deriving Bayesian posteriors is useful even when they are intractable since they can guide us towards better approximations.

As a first step, note that if $P$ is computable (or just lower semi-computable), then there exists an $m$ with $Q_m = P$. If $\pi_n$ is our prior belief that $\x$ was generated by $Q_n$, then
\begin{align}
    -U(\x) &= \log P(\x) - \log S(\x) \\
    &= \log \frac{\pi_m Q_m(\x)}{\sum_n \pi_n Q_n(\x)} - \log \pi_m \\
    &= \log \text{Pr}(m \mid \x) - \log \pi_m
\end{align}
can be seen as the log-posterior probability of $P$ given $\x$ up to a constant, consistent with our earlier notion of realism.% (Section~\ref{sec:typicality}).

\subsection{Batched Universal Critics}
\label{sec:batched}

How does our new notion of realism compare to existing notions of realism? $U$ is a particular instance of a no-reference metric since it can be applied to a single instance $\x$. But it turns out that we can also use it to approximate divergences by taking averages, as we will demonstrate. Consider evaluating the distribution $Q$ based on its average realism score as assigned by $U$. We have
\begin{align}
    \mathbb{E}_Q[U(\x)]
    &= \mathbb{E}_Q[\log S(\x) - \log P(\x)] \\
    &\leq \mathbb{E}_Q[\log Q(\x) - \log P(\x)] \label{eq:upper_bound} \\
    &= \KLD{Q}{P},
\end{align}
where Eq.~\ref{eq:upper_bound} is due to $Q$ minimizing cross-entropy when the data is distributed according to $Q$. On the other hand, if $Q$ is computable (or just lower semicomputable), we have
\begin{align}
    \mathbb{E}_Q[U(\x)]
    &= \textstyle \mathbb{E}_Q[\log \sum_n \pi_n Q_n(\x) - \log P(\x)] \\
    &\geq \mathbb{E}_Q[\log (\pi_m Q_m(\x)) - \log P(\x)] \label{eq:lower_bound} \\
    &= \textstyle \KLD{Q}{P} - \log \frac{1}{\pi_Q}
\end{align}
since we must have $Q_m = Q$ for some $m$. For ease of notation, we also write $\pi_Q$ to refer to $\pi_m$. The inequality follows because the terms we dropped from the sum are all non-negative. What this sandwich bound implies is that our universal critic works well as a replacement for the optimal critic $T_Q$ (Eq.~\ref{eq:optimal_critic}) if the \textit{complexity of $Q$}, $\log (1/\pi_Q)$, is low relative to the KL divergence between $Q$ and $P$. This agrees with our intuition for realism. In particular, we are more likely to accept an alternative explanation of the data if the explanation is simple, that is, if it can be described in a few words (or bits). A sequence of zeros (Examples~1 and 2) is easy to detect because it is cheap to describe (``always output 0'').
While the critic $U$ depends on $P$, it is \textit{universal} in the sense that it does not depend on $Q$.

\begin{tcolorbox}
\textbf{Example 4 (Low Complexity).} Consider a distribution over natural images $P$ and a distribution $Q_0$ which assigns all its mass to a single flat image, $Q_0(\x = 0) = 1$. Based on our bounds above, we should expect $U$ to detect $Q_0$ as unrealistic since it is cheap to describe, that is, $\log (1/\pi_{Q_0})$ is small for any reasonable coding scheme. In contrast, using $-\log P(\x)$ instead of $U(\x)$ would fail to detect $Q_0$ since natural image distributions generally assign high probability to flat images.
Similarly, images of Gaussian white noise would be detected since their distribution is cheap to describe as independent copies of a simple distribution.
\end{tcolorbox}

Note from Example~4 that low-complexity distributions can have both low or high entropy, that is, the complexity (or coding cost) $\log(1/\pi_Q)$ of a distribution $Q$ is different from its entropy.

\begin{tcolorbox}
\textbf{Example 5 (High Complexity).} As another example, consider a distribution which has memorized a training set of natural images, $Q_\mathcal{D}(\x) \propto \sum_{\x' \in \mathcal{D}} \delta_{\x'}(\x).$ This distribution will remain undetected since its complexity is high. To describe $Q_\mathcal{D}$, we would have to encode every image in the training set $\mathcal{D}$. On the one hand, this means that $U$ may perform poorly as an approximation of the KL divergence between $Q_\mathcal{D}$ and $P$ (due to the loose lower bound, Eq.~\ref{eq:lower_bound}). On the other hand, this behavior is in line with our intuitive notion of realism since we would also fail to tell a single example generated by $P$ from a single example selected from the training set. Like the universal critic, we consider training set images to be realistic\footnotemark.
\end{tcolorbox}

As a side note, a tighter bound can be obtained by choosing $m$ which maximizes $\pi_m Q_m(\x)$ instead of choosing $m$ with $Q_m = Q$ as in Eq.~\ref{eq:lower_bound}. This would correspond to the \textit{minimum description length} (MDL) principle of selecting models based on the total cost of describing the data and the model \citep{rissanen1978mdl}. That is, where adversarial training uses objectives such as maximum likelihood to select a critic, the universal critic can be viewed as selecting a critic based on MDL.

\footnotetext{More concretely, we can say that an average training set image would be considered realistic in the sense that $\mathbb{E}_\mathcal{D}[\mathbb{E}_{Q_\mathcal{D}}[U(\x)]] = \mathbb{E}_P[U(\x)] \leq 0$ (Eq.~\ref{eq:upper_bound}).}

We can further improve the critic's odds of detecting $Q$ by feeding it multiple independent examples. We define a \textit{batched universal critic} as a critic of the form
\begin{align}
    U^B(\x^B) = \log \sum_n \pi_n \prod_b Q_n(\x_b) - \log \prod_b P(\x_b),
    \label{eq:buc}
\end{align}
where $\x^B = (\x_1, \dots, \x_B)$. In the following, let $Q^B$ indicate the product measure, that is, a distribution over $B$ independent samples from $Q$. Then 
\begin{align}
   &\textstyle\frac{1}{B} \mathbb{E}_{Q^B}[U^B(\x^B)] \label{eq:buc_bound_1} \\
   \geq\ &\textstyle\frac{1}{B} \mathbb{E}_{Q^B}\left[ \log \left(\pi_m Q_m^B(\x^B)\right) - \log P^B(\x^B) \right] \\
   =\ &\textstyle\frac{1}{B}\sum_b \mathbb{E}_{Q}\left[ \log Q_m(\x_b) - \log P(\x_b) \right] + \frac{1}{B} \log \pi_m \notag \\
   =\ &\textstyle\mathbb{E}_{Q}\left[ \log Q(\x_b) - \log P(\x_b) \right] + \frac{1}{B} \log \pi_Q \\
   =\ &\textstyle\KLD{Q}{P} - \frac{1}{B} \log \frac{1}{\pi_Q}
   \label{eq:buc_bound}
\end{align}
for some $m$ where $Q_m = Q$. 
Compared to Eq.~\ref{eq:lower_bound}, we now obtain a tighter bound, which agrees with our intuition that upon observing multiple examples we should be able to do a better job of discriminating $Q$ from $P$. 
In the limit of large $B$ we recover the KL divergence. In this sense our notion of realism generalizes prior notions of realism based on no-reference metrics or divergences, and allows us to interpolate between the two.

\subsection{Universal Tests}
\label{sec:universal_test}

Deciding whether $\x$ is realistic or not means deciding between two hypotheses. The null hypothesis is that $\x$ is realistic, by which we mean that $\x$ came about in a particular way, modelled by $\x$ being drawn from the distribution $P$. Our alternative hypothesis is that $\x$ is unrealistic, or that it came about by some other process $Q$. For example, $P$ may be a distribution over photos but an alternative explanation could involve heavy compression with JPEG, corresponding to a distribution over images with blocking artefacts. If there are multiple ways in which $\x$ can fail to be realistic, $Q_n$, then it is natural to assign probabilities $\pi_n$ to these events and to consider a mixture distribution as our alternative hypothesis. We end up with $S$ as our alternative hypothesis if the only assumption we are willing to make is that $\x$ was generated by some computable process. By the well-known \textit{Neyman-Pearson lemma} \citep{neyman1933}, the most powerful test is then a likelihood ratio test of the form
\begin{align}
    \log S(\x) - \log P(\x) > \eta,
\end{align}
where $\eta$ is a parameter which controls the trade-off between false positives and false negatives. Note that the left-hand side is our universal critic. If we accept the Neyman-Pearson lemma then it is easy to accept that our measure of realism should take the form of a likelihood ratio instead of just $P(\x)$. However, this does not yet explain why our choice of alternative hypothesis should be $S$.
 
We can provide the following additional justification for the universal critic. Assume that instead of $S$ we decide to use another alternative hypothesis corresponding to a computable (or just lower semicomputable) measure $Q$. Then it is not difficult to see that
\begin{align}
    \textstyle U^B(\x^B) \geq \log Q^B(\x^B) - \log P^B(\x^B) - \log \frac{1}{\pi_Q}
\end{align}
for all $\x^B$ and all $B$ (following the same reasoning as in Eqs.~\ref{eq:buc_bound_1}-\ref{eq:buc_bound}). That is, $U^B$ additively dominates any computable likelihood ratio test and the constant $\log(1/\pi_Q)$ becomes negligible for sufficiently large $B$. Asymptotically, the universal critic is as sensitive to unrealistic examples as any other test based on an alternative hypothesis $Q$.\footnote{\citet[Chapter 4.3]{li1997intro} proved the stronger result that randomness deficiency additively dominates any so-called sum-$P$ test.}

\subsection{MCMC}
\label{sec:mcmc}

When optimizing data for realism it is natural to look to Markov chain Monte Carlo (MCMC) methods for solutions. In MCMC, the data is stochastically perturbed until it converges to a sample from our target distribution $P$ (at which point it would appear realistic). For example, for a continuous distribution with differentiable density $p$, a simple MCMC strategy based on \textit{Langevin diffusion} uses updates of the form
\begin{align}
    \textstyle \x_{t + \varepsilon} = \x_t + \varepsilon \left(\nabla \log p(\x_t) + \sqrt{2} \bm{\eta}_t \right),
    \label{eq:diffusion}
\end{align}
where $\bm{\eta}_t \sim \mathcal{N}(0, \mathbf{I})$ is independent Gaussian noise. For infinitesimal $\varepsilon$, the sequence of $\x_t$ converges to the distribution $P$. For a fixed $\varepsilon > 0$ the stationary distribution will only approximate $P$, but this can be addressed by performing additional Metropolis-Hastings accept/reject steps \citep{besag1994mcmc,welling2011mala}.

While MCMC produces realistic examples, it is not directly applicable to problems of the form of Eq.~\ref{eq:rdp}, since it is unclear how to translate an MCMC algorithm into a loss function $U$. If we naively interpreted Eq.~\ref{eq:diffusion} as a noisy gradient update, then this would correspond to using $p$ as a measure of realism and is bound to fail (Section~\ref{sec:probability}).

In a second attempt to make MCMC work for us, consider the sequence of distributions generated by Eq.~\ref{eq:diffusion}. Let $q^0$ be the density used to initialize~$\x_0$. Then each update produces a new density $q^t$ which approaches $p$ as $t$ goes to infinity. \citet{maoutsa2020fp} and \citet{song2021scorebased} showed that the deterministic updates
\begin{align}
    \x_{t + \varepsilon} = \x_t + \varepsilon \left( \nabla \log p(\x_t)  - \nabla \log q^t(\x_t) \right)
    \label{eq:diff_deterministic}
\end{align}
follow the same sequence of distributions $q^t$ (for infinitesimal $\varepsilon$, or approximately for $\varepsilon > 0$). Eq.~\ref{eq:diff_deterministic} suggests moving $\x_t$ towards high-density regions of $p$ but away from high-density regions of its \textit{current distribution} $q^t$. When optimizing for realism, we do not know $q^t$. But assuming an underlying $q^t$ exists, a Bayesian approach would be to estimate the missing gradient in Eq.~\ref{eq:diff_deterministic} by assigning prior probabilities $\pi_n$ to candidate densities $q_n$ and then to form the posterior expectation
\begin{align}
   \textstyle\sum_n P(n \mid \mathbf{x}_t) \nabla \log q_n(\x_t)
    &= \textstyle \nabla \log \sum_n \pi_n q_n(\x_t)
\end{align}
where $P(n \mid \mathbf{x}_t) \propto \pi_n q_n(\x_t)$ (Appendix~\ref{app:mcmc}). Note the resemblance of the right-hand side to Solomonoff's probability. If we restrict the universal critic to distributions with differentiable densities, then gradient descent on its density can be viewed as a Bayesian's attempt to simulate Eq.~\ref{eq:diff_deterministic}.

\subsection{Limited-Memory Observer}
\label{sec:observer}

We demonstrated useful statistical properties of universal critics and discussed connections to adversarial critics, significance testing, and MCMC. However, did we capture anything about \textit{how humans perceive inputs}? In this section we will argue that batched universal critics not only generalize no-reference metrics and divergences, but also represent a more realistic model of human observers. 

No-reference metrics are motivated by the idea that humans can look at a single image and decide whether it is realistic or not. It should therefore be possible to design a function which performs this task similarly well. However, in practice, even human observers often have access to not just a single image but a number of images. When evaluating the quality of image codecs or generative models, for example, human raters typically receive a stream of images and are asked to rate them. Mean opinion score tests ask raters to assign a score between 1 and 5 to each image while an alternative approach asks raters to classify between real and generated images \citep{denton2015gan}.
A generative model which always produces the same output would easily be identified by humans in such a task, even when the image appears realistic when viewed in isolation. While humans would be able to better detect a faulty generative model over time, no-reference metrics continue to produce the same output no matter how many examples they receive. That is, a no-reference metric is memoryless. While it may have been obtained through training on a set of realistic and unrealistic examples, it is unable to adapt to the method(s) currently under evaluation once it has been fixed.

Divergences represent the other extreme as they have access to the entire distribution. This corresponds to a human observer who has received an infinite stream of examples of either real or generated data. The total variation distance, for example, measures the probability of an \textit{optimal observer} correctly classifying real from generated data \citep{nguyen2009fdiv,blau2018tradeoff},
\begin{align}
    p_\text{success} = \frac{1}{2} D_\text{TV}[Q, P] + \frac{1}{2},
\end{align}
that is, an observer who has had access to infinitely many training examples. Other divergences can be similarly interpreted as classifiers which are optimal but with respect to different losses \citep{nguyen2009fdiv}. 

Like other no-reference metrics and human observers, universal critics provide a score for individual examples. Like divergences they can also be viewed as the score of a classifier deciding between two hypotheses, but unlike divergences they only have access to a finite set of training examples. This limitation means that prior assumptions become more important. Alternatively, universal critics can be viewed as measuring the performance of an ideal observer with limited memory (Appendix~\ref{app:observer}).
In this sense, batched universal critics are a better model of human observers than either no-reference metrics (memoryless) or divergences (infinite memory).

Universal critics as defined in Eq.~\ref{eq:uc} depend on an uncomputable Kolmogorov complexity and therefore could be implemented neither by humans nor computers. Given sufficient evidence, it will detect any failures a human observer might detect (Section~\ref{sec:universal_test}) but will also detect any unrealistic properties that would be missed by us. In this sense, universal critics provide a sufficient but not necessary criterion for high perceptual quality (unlike typicality, which is necessary but not sufficient).
The limitations of human observers can be incorporated naturally into universal critics by limiting $S$ to a mixture over fewer components. However, characterizing the limitations and abilities of human observers is beyond the scope of this paper. We refer to \citet{griffiths2003algo}, who studied the ability of humans to detect randomness in binary sequences, and compared it to algorithmic notions of randomness.

\section{Related Work}
\label{sec:related}

Given the wide range of related fields and the vast amount of work in them (Section~\ref{sec:intro}), it is impossible to review any meaningful fraction of related work here. Instead, we will focus on two successful examples with interesting connections to universal critics.

\subsection{Input Complexity}

Several papers on outlier detection made the puzzling observation that generative models trained on one dataset of images can assign higher probability to other datasets \citep{choi2019waic,nalisnick2019do,hendrycks2019deep}. \citet{serra2020ood} found that the issue virtually disappears if instead of measuring log-probabilities, the negative log-probability under the model is compared with the coding cost of a lossless image compression method such as PNG,
\begin{align}
    -\log P(\x) - C(\x),
\end{align}
where $C(\x)$ is the coding cost obtained via compression. The authors found that this signal performed significantly better for outlier detection, providing support for our definition of realism (Eq.~\ref{eq:uc}) by viewing $C(\x)$ as an approximation to Kolmogorov complexity. It is further enouraging that a simple but flexible compression scheme can provide a useful signal. An interesting question for future research is what a differentiable analogue of $C$ would look like, and whether it can be made robust enough for optimization.

We note that input complexity has also been considered in statistics for its applications in hypothesis testing, including as an approximation to universal tests \citep{ryabko2006stats}.

\subsection{Score Distillation Sampling}

Score distillation sampling \citep[SDS;][]{poole2023df} is a technique which has gained a lot of popularity for training 3D generative models. Training 3D generative models is challenging due to the high cost associated with collecting 3D data. SDS tries to overcome these limitations by leveraging diffusion models \citep{sohldickstein15diff} trained on large amounts of 2D images to guide text-to-3D models towards realistic outputs. Briefly, diffusion models define latent variables $\z_t = \alpha_t \x + \sigma_t \bm{\epsilon}$ where $\bm{\epsilon} \sim \mathcal{N}(0, \mathbf{I})$ and a function $\hat\epsilon_t(\z_t)$ is trained to predict $\bm{\epsilon}$. For a conditional diffusion model whose outputs depend on text $y$, we have the important relationship \citep{robbins1956}
\begin{align}
    \hat\epsilon_t(\z_t; y) \approx \mathbb{E}[\bm{\epsilon} \mid \z_t, y] = -\sigma_t \nabla_{\z_t} \log p_t(\z_t \mid y),
    \label{eq:cond_exp}
\end{align}
where $p_t$ is the distribution of $\z_t$ so that $\hat\epsilon_t$ can also be used to estimate the gradient of these log-densities.

Simplifying a bit, \citet[SDS;][]{poole2023df} propose the following gradient,
\begin{align}
    \nabla_{\x} \mathcal{L}_\text{SDS}(\x; y)
    &= \mathbb{E}_{t,\bm{\epsilon}}[w(t) (\hat\epsilon_t(\z_t; y) - \bm{\epsilon}) ]
\end{align}
where $w(t)$ are hyperparameters assigning weights to the different noise levels.
Is $\mathcal{L}_\text{SDS}$ a good candidate for $U$? We can see that SDS tries to find $\x$ such that $\z_t$ is near modes of $p_t$.
Note that $p_t$ is essentially the density of $\x$ smoothed via convolution with a Gaussian kernel, and so SDS appears fundamentally similar to using $p$ as a measure of realism (Section~\ref{sec:probability}) and susceptible to similar failures. Indeed, if the data distribution was Gaussian, then $p_t$ would also be Gaussian and the optimal $\x$ would be the mean, which tends to be unrealistic. This raises the question of why SDS performs well in practice. The key to its success lies in classifier-free guidance \citep[CFG;][]{ho2021cfg}. Instead of using $\hat\epsilon_t$ directly, this common trick is to use
\begin{align}
    \hat\epsilon_t^v(\z_t; y) = (1 + v) \hat\epsilon_t(\z_t; y) - v \hat\epsilon_t(\z_t),
    \label{eq:cfg1}
\end{align}
where $\hat\epsilon_t(\z_t)$ is an unconditional prediction of $\bm{\epsilon}$ and the guidance weight $v \geq 0$ is a hyperparemeter. This corresponds to a gradient signal proportional to
\begin{align}
     v \nabla_{\z_t} \log p_t(\z_t)
    - (1 + v) \nabla_{\z_t} \log p_t(\z_t \mid y).
    \label{eq:cfg2}
\end{align}
Implicit in the marginal density $p_t(\z_t)$ is a large mixture over all possible texts $y$,
\begin{align}
    \textstyle p_t(\z_t) = \sum_y p(y) p_t(\z_t \mid y).
\end{align}
Note the resemblance of Eq.~\ref{eq:cfg2} to our universal critic. For large $v$, the constant $1$ becomes negligible and we are left with a density ratio between the target distribution and a large mixture distribution over alternative explanations. Indeed, \citet{poole2023df} found that SDS without CFG produced blurry 3D scenes and very large guidance weights worked best.

We therefore submit that the reason SDS works well is not explained by its ability to find modes in densities or its connections to model distillation techniques, but by its ability to approximate universal critics. Reinterpreting SDS in this way suggests new ways of overcoming its weaknesses (e.g., its tendency to produce oversaturated images), such as a more intentional design of the mixture of alternatives, or batched losses analogous to Eq.~\ref{eq:buc}. 

\section{Discussion}

In this position paper we have argued that the question of realism is equivalent to the question of randomness, that is, whether observations originated from a particular distribution. This allowed us to draw on insights from algorithmic information theory and to propose universal critics, or randomness deficiency \citep{li1997intro}, as a rational answer.
Perceptual quality can be viewed as the result of a (necessarily) imperfect approximation of universal critics. However, despite the relevance of these concepts to problems in machine learning, discussions of randomness deficiency are notably absent from its literature. Instead, dominant notions of realism continue to be based on probability \citep[e.g.,][]{ruff2021anomaly,poole2023df}, typicality \citep[e.g.,][]{nalisnick2019typicality} or divergences \citep[e.g.,][]{blau2018tradeoff,theis2021coding}.

A divergence of zero is a sufficient condition for perfect realism but corresponds to an ideal observer with access to an infinite stream of examples. As such, it is stronger than required for most practical applications where observers only have access to one or a few examples. At the other end of the spectrum, weak typicality is an example of a criterion which only considers a necessary criterion, while most no-reference metrics correspond to neither a necessary nor a sufficient criterion (e.g., high probability in some feature space). Universal critics enable principled relaxations of divergence-based constraints. While weaker than divergences (in the desired way), they are still strong in the sense that they are as strong as other likelihood-ratio tests for realism, up to a constant which depends on the complexity of the competing test (Section~\ref{sec:universal_test}).

Many interesting practical and theoretical questions remain. For example, what is the impact of different choices of $\pi_n$ on the sample efficiency of universal critics? What are the implications of using universal critics in rate-distortion-realism trade-offs? Most importantly, what do practical approximations to universal critics (Eqs.~\ref{eq:uc} or \ref{eq:buc}) look like that can serve as optimization targets?

\section*{Acknowledgements}

I would like to thank Aaron B. Wagner and Johannes Ballé for many helpful discussions shaping the arguments presented in this paper, Andriy Mnih, Matthias Bauer, Jörg Bornschein, Iryna Korshunova, Emilien Dupont, Eirikur Agustsson, and Alexandre Galashov for valuable feedback on the manuscript, and Daniel Severo for exploring various ideas to make universal critics practical.

\bibliographystyle{plainnat}
\bibliography{references}

\newpage
\appendix
\onecolumn

\section{Typicality}
\label{app:typicality}

 In Appendix~\ref{app:bounded_size} we extend our discussion of weak typicality. In particular, we elaborate on how a majority of examples in the typical set can be unrealistic. 
 In Appendix~\ref{app:strong_typicality} we additionally consider \textit{strong typicality}. 
 
\subsection{Bounded size of weakly typical sets}
\label{app:bounded_size}

As discussed in the main text, the typical set contains sequences $\x^N \sim P^N$ with high probability, that is, $A_\delta^N$ is large enough that $P^N(A_\delta^N)$ approaches 1 as $N$ goes to infinity. On the other hand, the typical set is small in the sense that the number of elements is bounded by \citep{cover2006elements}
\begin{align}
    |A_\delta^N| \leq 2^{H[\x^N] + N\delta}.
    \label{eq:typical_set_size}
\end{align}
This fact is exploited in information theory to build simple but efficient codes for data compression. Using
\begin{align}
    \log_2 |A_\delta^N| + 1 \leq H[\x^N] + N\delta + 1
\end{align}
bits, we can address each element in the typical set. Normalized by the number of elements, this becomes
\begin{align}
    \frac{1}{N} H[\x^N] + \delta + \frac{1}{N} = 
    H[\x_n] + \delta + \frac{1}{N},
    \label{eq:cc_overhead}
\end{align}
approaching the entropy of $P$ as $N$ increases and $\delta$ decreases. Counter-intuitively, this suggests that the typical set cannot contain too many unrealistic sequences, or else our compression scheme would be inefficient. However, note that while the coding rate overhead is only $(\delta + 1/N)$ above $H[\x_n]$ (Eq.~\ref{eq:cc_overhead}), the number of elements in the typical set already exceeds $2^{H[\x]}$ by a factor of up to $2^{N\delta}$ (Eq.~\ref{eq:typical_set_size}). If we relax the threshold $\delta$ so that $N\delta$ increases by 1, then this would increase the total coding cost of a sequence by only 1 bit, yet the number of elements in the typical set increases by a factor of up to 2.

\subsection{Strong typicality}
\label{app:strong_typicality}

Here we consider strong typicality. A closely related notion ($P$-typicality) was considered by \citet{chen2022rdpf} to quantify the realism of a batch of examples.
Strong typicality is also similar in spirit to maximum mean discrepancy \citep[MMD;][]{gretton2012mmd}, which was discussed in Section~\ref{sec:mmd}. 

Let $\mathcal{X}$ be a finite set and let $\#(x, \x^N)$ be the number of occurrences of $x$ in a sequence $\x^N = (x_1, \dots, x_N)$, that is, a histogram. The set of \textit{strongly typical} sequences is defined as \citep{cover2006elements}
\begin{align}
    T_\delta^N = \left\{ \x^N \in \mathcal{X}^N : \sum_{x \in \mathcal{X}} \left|\frac{1}{N} \#(x, \x^N) - P(x) \right| < \delta \right\}.
\end{align}
As for weakly typical sets $A_\delta^N$, the probability that a randomly drawn sequence $\x^N \sim P^N$ is strongly typical approaches 1 for any $\delta > 0$ as $N$ increases. Strong typicality requires the empirical distribution of elements in a sequence to be close to the distribution of interest, $P$. For large $N$, a randomly selected element of a strongly typical sequence will appear like a sample from $P$, that is, it will appear realistic. However, the main challenge we are trying to overcome is to define realism for short sequences and individual $x$. If we naively apply strong typicality to a single element ($N = 1$), we obtain
\begin{align}
    \textstyle\sum_{x \in \mathcal{X}} \left|\#(x, (x_1)) - P(x) \right|
    &= |1 - P(x_1)| + \textstyle\sum_{x \neq x_1} |0 - P(x)| \\
    &= 1 - P(x_1) +  \textstyle\sum_{x \neq x_1} P(x) \\
    &= 1 - P(x_1) + 1 - P(x_1) \\
    &= 2 - 2 P(x_1),
\end{align}
that is, we are effectively back to measuring the probability of $x_1$. It is therefore unclear how strong typicality could be used to evaluate objects as high-dimensional as images. One might consider dividing an image into patches of lower dimensionality and treating the image as a sequence of these. However, this would ignore dependencies between patches and we would further have to assume that the statistics of each realistic image is representative of the entire distribution (i.e., ergodicity), which may not be the case.

\section{Expected gradient of log-density}
\label{app:mcmc}

Let $P(n \mid \x) \propto \pi_n q_n(\x)$ be the posterior probability that $\x$ was drawn from $q_n$. Then the expected gradient of the log-density is:
\begin{align}
    \textstyle\sum_n P(n \mid \mathbf{x}) \nabla \log q_n(\x) 
    &= \textstyle\sum_n P(n \mid \mathbf{x}) \frac{1}{q_n(\x)} \nabla q_n(\x) \\
    &= \textstyle\sum_n \frac{\pi_n q_n(\mathbf{x})}{\sum_m \pi_m q_m(\mathbf{x})} \frac{1}{q_n(\x)} \nabla q_n(\x) \\
    &= \textstyle\frac{1}{\sum_n \pi_n q(\mathbf{x})} \nabla \sum_n \pi_n q_n(\x) \\
    &= \nabla \log \sum_n \pi_n q_n(\x)
\end{align}

\section{Limited-memory observer}
\label{app:observer}

Here we elaborate on the relationship between the batched universal critic and an ideal observer in a sequential prediction task. Assume an observer assigns a value $T(\x)$ to an image $\x$. Further assume we ask the observer to
\begin{align}
    \underset{T}{\text{maximize}} \quad \mathbb{E}_Q[T(\x)] - \mathbb{E}_P[\exp(T(\x))],
\end{align}
that is, the observer receives a reward of $T(\x)$ if $\x \sim Q$ and a penalty of  $\exp(T(\x))$ if $\x \sim P$.
The optimal output is then given by \citep{glaser2021kale}
\begin{align}
    T_Q(\x) &= \log Q(\x) - \log P(\x).
\end{align}

Note that
\begin{align}
    \mathbb{E}_P[\exp(T_Q(\x))] = \mathbb{E}_P[Q(\x) / P(\x)] = \sum_{\x} Q(\x) = 1.
\end{align}
$T_Q$ remains the optimal solution if we solve the following closely related constrained optimization problem,
\begin{align}
    \underset{T}{\text{maximize}} \quad \mathbb{E}_Q[T(\x)]  \quad\text{subject to}\quad \mathbb{E}_P[\exp(T(\x))] \leq 1,
\end{align}
and so we can use $\mathbb{E}_Q[T(\x)]$ to evaluate $T$ if we fix $P$ and restrict the class of allowed $T$ in this way. In other words, an equivalent task presents raters only with examples from $Q$, but applies restrictions to the scores that can be assigned.

Unlike typical classification problems where both $P$ and $Q$ are unknown and must be learned, we can assume $P$ to be known to the observer through prior experience while $Q$ still needs to be learned.
A rational observer who expects $\x$ to be distributed according $Q_n$ with probability $\pi_n$ would maximize the expected reward by using
\begin{align}
    U(\x) = \log P(\x) - \log S(\x), \quad\text{where}\quad S(\x) = \textstyle\sum_n \pi_n Q_n(\x).
\end{align}
After receiving $B$ examples from $Q$, $\x^B = (\x_1, \dots, \x_B)$, a rational observer would update those beliefs to
\begin{align}
    \pi(n \mid \x^B) \propto \pi_n Q_n^B(\x^B) = \pi_n \textstyle\prod_{b = 1}^B Q_n(\x_b)
\end{align}
and receive a reward of
\begin{align}
    U(\x \mid \x^B) = \log P(\x) - \log S(\x \mid \x^B), \quad\text{where}\quad S(\x \mid \x^B) = \textstyle\sum_n \pi(n \mid \x^B) Q_n(\x)
\end{align}
for a subsequent example $\x$ from the unknown $Q$. This score is slightly different from the batched universal critic, which is more readily interpreted as the combined value assigned to an entire batch of examples. However, the following relationship holds:
\begin{align}
    U(\x^B)
    =\ &\log P^B(\x^B) - \log \textstyle\sum_n \pi_n Q_n^B(\x^B) \\
    =\ &\log P^B(\x^B) - \log \textstyle\sum_n \pi_n Q_n^{B - 1}(\x^{B - 1}) Q_n(\x_B) \\
    =\ &\log P^B(\x^B) - \log \textstyle\sum_n \frac{\pi_n Q_n^{B - 1}(\x^{B - 1})}{\sum_m \pi_m Q_m^{B - 1}(\x^{B - 1})} Q_n(\x_B) - \log \sum_m \pi_m Q_m^{B - 1}(\x^{B - 1}) \\
    =\ &\log P(\x_B) - \log \textstyle\sum_n \pi(n \mid \x^{B - 1}) Q_n(\x_B) + \log P^{B - 1}(\x^{B - 1}) - \log \sum_m \pi_m Q_m^{B - 1}(\x^{B - 1}) \\
    =\ &U(\x^B \mid \x^{B - 1}) + U(\x^{B - 1}) \\
    =\ &\textstyle\sum_{b = 1}^B U(\x^b \mid \x^{b - 1})
\end{align}
That is, the output of the batched universal critic can be viewed as the sum of scores achieved in $B$ sequential prediction tasks.

\end{document}